# A study of the singularity locus in the joint space of planar parallel manipulators: special focus on cusps and nodes

Abdel Kader Zaiter, Philippe Wenger and Damien Chablat

L'Institut de Recherche en Communications et Cybernétique de Nantes (*IRCCyN*), École Central de Nantes, 1 rue de la Noë, 44321 Nantes, Cedex 3, France. Emails: {Abdel-Kader.Zaiter, Philippe.Wenger, Damien.Chablat}@irccyn.ec-nantes.fr}

*Résumé*— Dans une section de l'espace articulaire des manipulateurs parallèles, les points cusps et noeuds ont un rôle essentiel dans le changement de mode d'assemblage. Ce travail présente une étude détaillée de ces points dans l'espace articulaire et de travail. On démontre qu'un point cusp (resp. un noeud) définit un point de tangence (resp. un point d'intersection), dans l'espace de travail, entre les courbes singulières et les courbes associées aux surfaces caractéristiques. L'étude est illustrée sur le manipulateur planaire 3-*RPR* mais son champ d'application est général.

*Abstract*— **Cusps and nodes on plane sections of the singularity locus in the joint space of parallel manipulators play an important role in nonsingular assembly-mode changing motions. This paper analyses in detail such points, both in the joint space and in the workspace. It is shown that a cusp (resp. a node) defines a point of tangency (resp. a crossing point) in the workspace between the singular curves and the curves associated with the so-called characteristics surfaces. The study is conducted on a planar 3-*RPR* manipulator for illustrative purposes.**

## 1. INTRODUCTION

Most parallel manipulators have singularities that limit the motion of the moving platform. The most dangerous ones are the singularities associated with the direct kinematics, where two direct kinematic solutions (DKSs) or assembly modes (AM) coalesce. Indeed, approaching such a singularity results in large actuator torques or forces, and in a loss of stiffness. Planar parallel manipulators have received a lot of attention [1-4, 6, 7, 9, 10, 15-16] because of their relative simplicity with respect to their spatial counterparts. Moreover, studying the former may help understand the latter. Planar manipulators with three extensible leg rods, referred to as 3-*RPR* manipulators, have often been studied. Such manipulators may have up to six assembly modes (AM) [1]. The direct kinematics can be written in a polynomial of degree six. Moreover, the singularities coincide with the set of configurations where two direct kinematic solutions coincide. It was first pointed out that to move from one assembly mode to another, the manipulator should cross a singularity [2]. However, [3] showed, using numerical experiments, that this statement is not true in general. More precisely, this statement is only true under some special geometric conditions, such as similar base and mobile platforms [4]. Relying on geometric arguments, [5] conjectured that the workspace of 3-*RPR* parallel manipulator is divided into two singularity-free regions called aspect and that there should be 3 solutions in each aspect. Recently, [6] provided a mathematical proof of the decomposition of the workspace into two aspects using geometric properties of the singularity surfaces. Also, non-singular AM changing motions were described in a 3-D representation. McAree and Daniel [4] pointed out that a 3-*RPR* planar parallel manipulator can execute a non-singular change of assembly-mode if a point with triple direct kinematic solutions exists in the joint space. The authors established a condition for three direct kinematic solutions to coincide and showed that the encirclement of a cusp point is a sufficient condition for a non-singular AM change. This condition was exploited by [7], where authors provided an algorithm to detect the cusp points of a 3-*RPR* parallel manipulator in a section of the joint space, which corresponds to an input joint variable set to a constant value. Wenger and Chablat [8] investigated the question of whether a change of assembly-mode must occur or not when moving between two prescribed poses in the workspace. They defined the uniqueness domains in the workspace as the maximal regions associated with a unique assembly-mode and proposed a calculation scheme for 3-*RPR* planar parallel manipulators using octrees. They showed that up to three uniqueness domains exist in each singularity-free region. When the starting and goal poses are in the same singularity-free region but in two distinct uniqueness domains, a non-singular change of assembly-mode is necessary. However they did not investigate the kind of motion that arises when executing a non-singular change of assembly-mode. [9] analyzed the variation of the topology of singularity curves and the distribution of the cusp points from one section of the joint space to another. By calculating the image in the workspace of a trajectory that encircles a cusp point, [10] described all corresponding trajectories. Also, the authors introduced the notion of reduced configuration space for an aspect and used this notion to show which cusp points must be encircled to achieve a non singular AM change in the associated aspect. [11] showed the singularity surface in the 3D workspace for the 3-RPR parallel manipulator, with the six solutions corresponding to the same point in the joint space. Moreover, they studied the coincidence of DKSs on the singular curves in the reduced configuration space. Another type of AM change was reported by [12], which corresponds to the encirclement of an α-curve in the joint space. Moreover, the authors showed the possibility to produce an AM change when approaching a singularity for micro mechanisms that present relatively large joint clearances. Later, [13] related the encirclement of a α-curve to the encirclement of a loop characterized by a node, which corresponds to the simultaneous coalescence of two couples of DKSs. [14] provided a mathematical condition of the existence of cusps and nodes. [15] provided a tool to calculate the practical workspace and, therefore, the reduced configuration space, keeping one input variable constant. This work was extended in [16] with more explanations and more examples. In [17], the authors showed, using numerical experiments, that not any cusp point may be encircled to perform a nonsingular AM changing motion. However, their work did not make it possible to identify definite rules.

In this work, a detailed analysis is provided to explain the role of the singular curves in the joint space and in the workspace as pertained to the loss of solutions. More attention is drawn to the study of cusps and nodes. It is shown that a cusp (resp. a node) defines a point of tangency (resp. a crossing point) in the workspace between the singular curves and the curves associated with the so-called characteristics surfaces. This study is illustrated with a *3-RPR* planar parallel manipulator. It is a first step towards a rigorous and complete analysis of nonsingular changing motions in parallel manipulators.

## 2. ILLUSTRATIVE MANIPULATOR

A *3-RPR* planar parallel manipulator is used in this paper to illustrate the analysis. As showed in figure 1, this manipulator has triangular base and platform denoted $A_1 A_2 A_3$ and $B_1 B_2 B_3$, respectively.

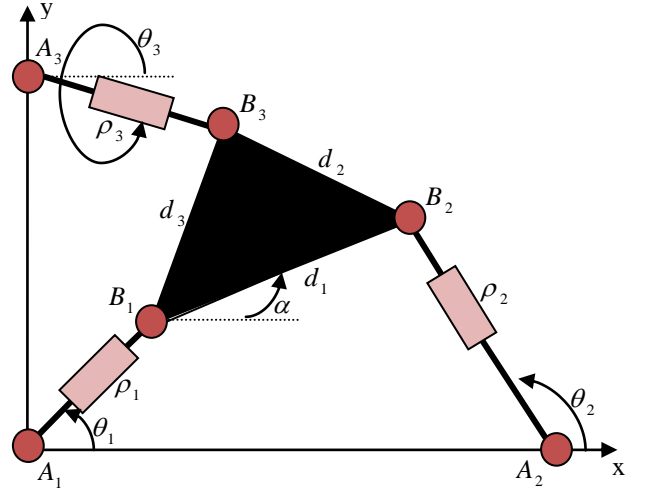

Figure 1. The 3-RPR planar parallel manipulator.

The base and the platform are related by three similar legs. Each leg consists of three joints; the first and third ones are passive revolute joints while the second one is an actuated prismatic joint. $\theta_i$, (i=1, 2 and 3) represents the passive variable that corresponds to the angles between the leg axis and the x-axis. The input variables are defined by $\rho_i$ (i=1, 2 and 3); the lengths of legs. The output variables are usually defined by the Cartesian coordinates $(x, y)$ of $B_1$ and $\alpha$; the orientation of the platform in the plane with respect to the x-axis. In this paper, the position coordinates will be more conveniently defined by ($\rho_1, \theta_1$) i.e., the cylindrical coordinates of $B_1$. With these parameters, indeed, it is possible to simplify the problem: we consider 2-dimensional slices of the joint space and of the workspace by fixing the joint parameter $\rho_1$. The projection of a ($\theta_1$, $\alpha$)-slice onto the ($\rho_2, \rho_3$) plane is consistent with the projection of the configuration space onto the joint space [4].

The geometric parameters of the illustrative manipulator are the same as those used in [3, 10, and 8], namely:
$d_1 = 17.04$, $d_2 = 16.54$, $d_3 = 20.84$, $A_1 = (0,0)$, $A_2 = (15.91, 0)$ and $A_3 = (0, 10)$.

## 3. SINGULARITIES
### 3.1 Singularities and aspects

The *3-RPR* parallel manipulator is in a singular configuration whenever the axes of its three legs are concurrent (the platform orientation gets out of control) or parallel (the translation along a direction orthogonal to the two parallel legs is out of control).

The singular configurations of the *3-RPR* planar parallel manipulator can be calculated without any difficulty. They define surfaces both in the workspace and in the joint space. The singular surfaces divide the workspace into two singularity-free domains called aspects [6, 5], referred to as $WA_1$ and $WA_2$. When mapped into the joint space, these two aspects coincide and thus define two coincident sets. Since we are considering slices by fixing $\rho_1$, the singularity locus can be depicted as curves in the workspace (resp. the joint space) in the ($\theta_1, \alpha$)-plane (resp. in the ($\rho_2, \rho_3$)-plane). At a point on a singularity curve in the workspace, the manipulator is necessarily in a singular configuration. At a point on a singularity curve that bounds the joint space, the manipulator is in a singular configuration. On an internal singularity curve, the point admits several nonsingular DKSs in addition to the singular solution. Near a singular point, there are two "mirrored" DKSs in the workspace on each part of the singular curve [3]. When a singular point of a singular curve is met in the joint space, the two mirrored DKSs coincide and one DKS is then lost in each aspect (see [3, 4] for example).

Plots of the singularity curves for $\rho_1 = 17$ are shown in figure 2. A *3-RPR* planar parallel manipulator has up to 6 assembly modes [1]. For a given point in the joint space that is not on a singular curve, there are 2, 4 or 6 DKSs in the workspace. The number of DKSs in each region of the joint space is indicated in figure 5.

### *3.2 Cusps and nodes*

Cusps and nodes appear on the singular curves as a consequence of the projection of the manipulator configuration space folds onto its joint space [4]. As described by Whitney [18], a cusp arises when a fold is "folded". Figure 3 shows the local model of a cusp where the singular locus is shown in red lines. Three DKSs coincide at a cusp point. A node appears on the singular curves in the joint space at a crossing point. A node arises from the projection of two distinct folds. There are two pairs of coincident DKSs at a node.

### *3.3 Characteristic surfaces, basic regions - basic components*

The notion of *characteristic surfaces* was introduced in [5]. Its definition is recalled thereafter.

Let $\partial WA_i$ define the boundary of aspect $WA_i$. The characteristic surfaces of $WA_i$, denoted by $S_c(WA_i)$, are defined as follows:

$$S_c(WA_i) = g^{-1}(g(\partial WA_i)) \cap WA_i \quad (1)$$

where:
$g$ maps all points of a given set of the workspace into the joint space through the manipulator inverse kinematics.
$g^{-1}$ maps all points of a given set of the joint space in the workspace through the manipulator direct kinematics.
Note that since we are considering slices by fixing $\rho_1$, the characteristic surfaces are defined by curves in this paper.
$S_c(WA_i)$ divides $WA_i$ into different *basic regions* $WAb_{ik}$,

$$\bigcup_{k \in K} WAb_{ik} = WA_i - S_c(WA_i) \quad (2)$$

where $K$ is the set indexing the basic regions.
The *basic components* $QAb_{ik}$ are defined as the images of $WAb_{ik}$ in the joint space:

$$QAb_{ik} = g(WAb_{ik}) \quad (3)$$

Since the *3-RPR* planar parallel manipulator has 2 aspects and up to 3 DKSs in each aspect, as many as 3 distinct points in each aspect map onto one unique point in the joint space. Thus, as many as 3 of the basic regions in the workspace map onto 3 coincident basic components in the jointspace. In figure 2a, the basic components are the domains bounded by parts of the singularity curves. Some of the basic regions are shown in figure 6.

The following notation will be used throughout the paper.
- $D_k$: direct kinematic solution number k. Let us label 1, 2, 3 the solutions in aspect $WA_1$ and 4, 5, 6 those in $WA_2$.
- $C_k$: curve segment on the singularity curves in the joint space where solution $D_k$ is lost when crossing $C_k$. Note that since one solution is lost in each of the two aspects, there will be always a mirrored solution $D_n$ in the second aspect that is lost simultaneously. Before disappearing, $D_n$ and $D_k$ coincide on $C_k$.
- $CP_{k-m}$: the cusp point that enables a nonsingular assembly-mode change between $D_k$ and $D_m$.
- $N_{k-m}$: node defined at the crossing point between $C_k$ and $C_m$.
- $Sc_k^m(i)$: the nonsingular curve image of $C_k$ in aspect $WA_i$ at the boundary of the basic region $WAb_{im}$.
- $S_k$: singular curve image of $C_k$ in the workspace.

Figure 2a shows the curve segments $C_k$ for our *3-RPR* planar parallel manipulator when $\rho_1 =17$.

## 4. SOLUTION LOSS ON THE SINGULARITY CURVES

Let take a first point $P_{initial}$ in a domain associated with 6 DKSs, 3 in each aspect. There are 3 coincident basic components associated with each aspect in this domain. Let take a second point $P_{final}$ in a neighboring domain with 4 DKSs. Point $P_{initial}$ (resp. $P_{final}$) is defined by $\rho_1 =17$, $\rho_2 =15$ and $\rho_3 =15$ (resp. $\rho_1 =17$, $\rho_2 =13.25$ and $\rho_3 =20.39$). Figure 2b shows why solutions $D_1$ and $D_5$ are lost when crossing $C_1 = C_5$ when going from $P_{initial}$ to $P_{final}$. We can determine the solution loss on all the other curve segments in a similar manner.

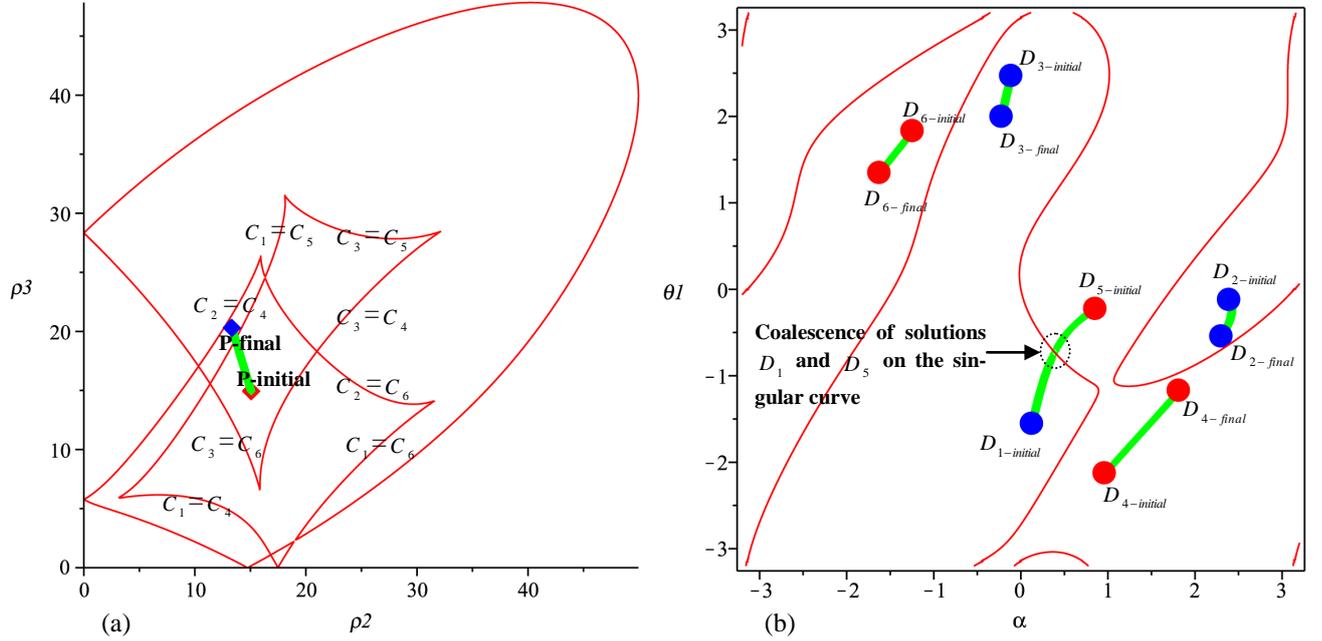

Figure 2. Loss of solutions when crossing singular curves in the joint space

## 5. CORRESPONDENCE OF CUSPS AND NODES IN THE WORKSPACE

In this section, it is shown that one image of a cusp point, which corresponds to the coalescence of three DKSs in the workspace, defines a tangency point between the curves associated with the characteristic surfaces and the singular curves in the workspace. Also, the other images will correspond to cusp points formed by the characteristic surfaces. For a node, two images will correspond to a crossing point between the curves associated with the characteristic surfaces and the singular curves in the workspace, where these two images correspond to the coalescence of two couples of DKSs. The other images of a node will correspond to cross points between curves associated to characteristic surfaces. For more simplicity, we consider a manipulator with two aspects, like the illustrative manipulator shown in figure 1.

### 5.1 Case of a cusp point

As shown in figure 4, a cusp point $CP_{k-m}$ appears where curve segments $C_k$ and $C_m$ meet. Both are parts of the boundary of coincident basic components. Since $CP_{k-m}$ allows a non singular AM change between solutions $D_k$ and $D_m$, these two solutions lie necessarily in the same aspect [4], say $WA_1$. In the workspace, $C_k$ defines one singular curve segment $S_k$ and one nonsingular curve segment $Sc_k^m(1)$ in aspect $WA_1$. $Sc_k^m(1)$ lies on the characteristic surface $S_c(WA_1)$ and defines a part of the boundary of the basic region $WAb_{1m}$. $S_k$ lies on the singular curve, on which $D_k$ is lost, and defines a part of the singular boundary of the basic region $WAb_{1k}$. Similarly, $C_m$ gives one non-singular curve segment $Sc_m^k(1)$ defining a part of the boundary of the basic region $WAb_{1k}$ in aspect $WA_1$ and one singular curve segment $S_m$ defining a part of the singular boundary of the basic region $WAb_{1m}$. As shown in figure 4, another solution $D_n$, that belongs to a second aspect $WA_2$, disappears at curves $C_k = C_n$ and $C_m = C_n$. Also, the singular curve segments $S_k = S_n$ and $S_m = S_n$, defined by $C_k$ and $C_m$, respectively, define two parts of the singular boundary of $WAb_{2n}$, where $WAb_{2n}$ lies in the second aspect $WA_2$. Since $CP_{k-m}$ belongs to both $C_k$ and $C_m$, its images in the workspace belong to $Sc_k^m(1)$, $Sc_m^k(1)$, $S_k$ and $S_m$. At $CP_{k-m}$, solutions $D_k$, $D_m$ and $D_n$ coalesce in the workspace, therefore this image of $CP_{k-m}$ is the point where the three basic regions $WAb_{1k}$, $WAb_{1m}$ and $WAb_{2n}$ meet in the workspace. Since $CP_{k-m}$ has one image $D_k$ in $WAb_{1k}$ having $Sc_m^k(1)$ and $S_k$ as parts of its boundary, where $CP_{k-m}$ has also an image on $Sc_m^k(1)$ and $S_k$, this image corresponds to the intersection point between $Sc_m^k(1)$ and $S_k$. By the same way, we can conclude that solution $D_m$ of $CP_{k-m}$ corresponds to the intersection point between $Sc_k^m(1)$ and $S_m$ at the boundary of $WAb_{1m}$, and $D_n$ corresponds to the intersection point between $S_k$ and $S_m$ at the boundary of $WAb_{2n}$. Consequently, the image of $CP_{k-m}$ that corresponds to the coalescence of $D_k$, $D_m$ and $D_n$ is the intersection point between $Sc_k^m(1)$, $Sc_m^k(1)$, $S_k$ and $S_m$ (see figure 4). From Whitney [18] and Corvez [19], the local model of a cusp always defines a point of tangency between the singular curve $C_f$ associated with f and the set $f^{-1}(f(C_f)) \setminus C_f$, as illustrated in figure 3. In our example, the sets $C_f$ and $f^{-1}(f(C_f)) \setminus C_f$ correspond to the curves $S_k \cup S_m$ and $Sc_k^m(1) \cup Sc_m^k(1)$, respectively. Thus $Sc_k^m(1) \cup Sc_m^k(1)$ is tangent to $S_k \cup S_m$ at the image of $CP_{k-m}$ in the workspace.

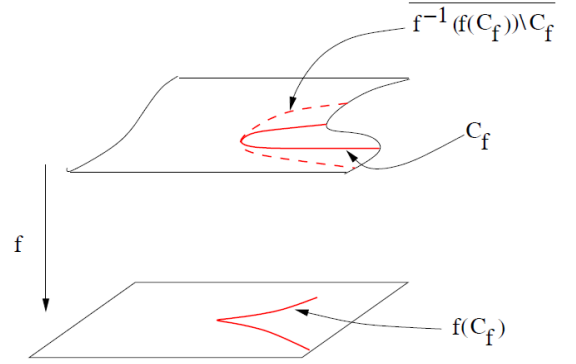

Figure 3. Tangency at a cusp point (from [19]).

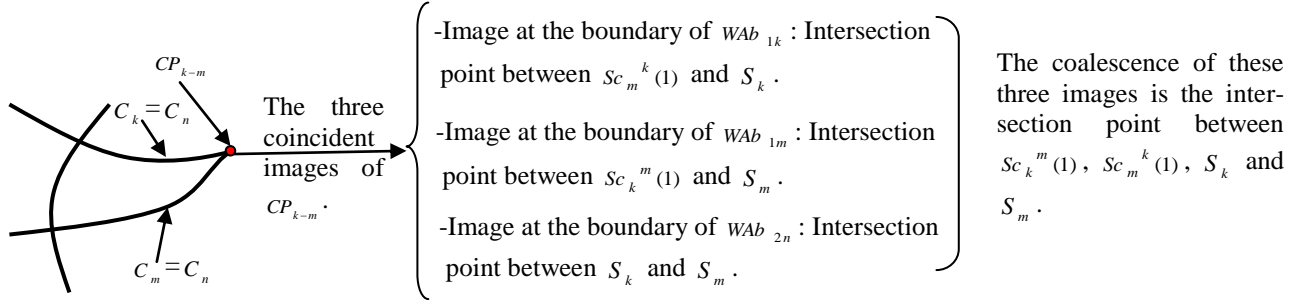

Figure 4. The correspondence of the three coincident solutions of CP$_{k-m}$.

### 5.2 Example on the cusp point

The 3-RPR planar parallel manipulator described in section 2 is chosen as illustrative example. We will explain the correspondence, in the workspace, of cusp point $CP_{4-6}$ shown in figure 5. This cusp point is formed by the parts of curves $C_4$ and $C_6$ colored in black and light blue, respectively. Near $CP_{4-6}$, each of these curves separates two domains in the joint space. The first domain corresponds to the coalescence of 3 basic components $QAb_{11} = QAb_{12} = QAb_{13}$ for aspect 1 plus 3 $QAb_{24} = QAb_{25} = QAb_{26}$ for aspect 2. In the second domain, 2 basic components coalesce (the number of DKSs is shown in figure 5). The 6 associated basic regions, which correspond to the first region, $WAb_{11}$, $WAb_{12}$ and $WAb_{13}$ in the first aspect and $WAb_{24}$, $WAb_{25}$ and $WAb_{26}$ in the second aspect, are shown in figure 6 and colored in dark green and orange, respectively. Any point on $C_4$ and $C_6$ yields 5 different solutions, one of which is a double solution that lies on the DKP singular curves. Therefore, the image of each one of the two curves yields 5 curve images in the workspace, where one curve image is a part of the singular curves and the four remaining images are parts of the characteristic surfaces. In figure 6, the curve images of curves $C_4$ and $C_6$ are represented by colors black and light blue, respectively. Each curve gives the expected number and types of curve images.

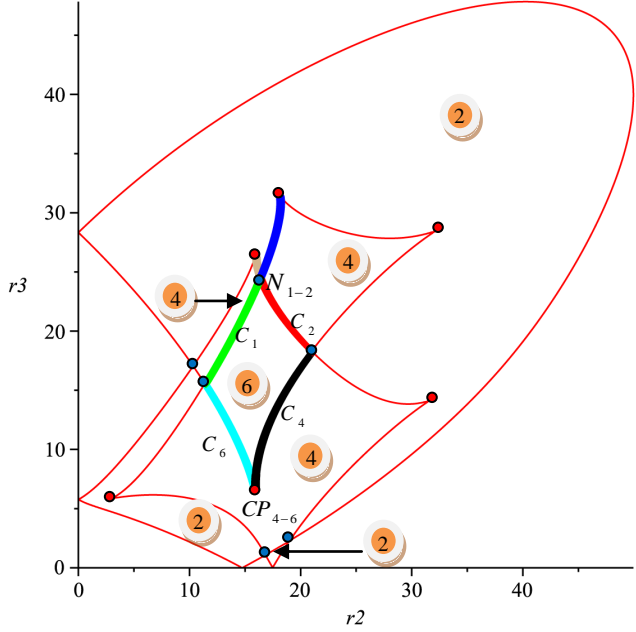

Figure 5. Cusp point $CP_{4-6}$ and node $N_{1-2}$ in the joint space. Number of DKSs in each region is indicated (orange disks)

Curve $C_4$ yields the curve image $Sc_4^6(2)$ at the boundary of $WAb_{26}$ and $S_4 = S_3$ at the common singular boundary of $WAb_{24}$ and $WAb_{13}$. Also, curve $C_6$ yields the curve image $Sc_6^4(2)$ at the boundary of $WAb_{24}$ and $S_6 = S_3$ at the common singular boundary of $WAb_{26}$ and $WAb_{13}$. Figure 6 shows that these four curve images have one intersection point that represents the image of $CP_{4-6}$ corresponding to the coalescence of three DKSs in the workspace. As expected, this image is a tangent point between the DKP singular curves and the characteristic surfaces.

It is known that, in the reduced configuration space, the different DKSs corresponding to same input variables represent the intersection points of a vertical line with the configurations space [10, 11]. For a cusp point, the vertical line is tangent on the singular curve folded in a specific way, where three DKSs coalesce. The projection of this fold on the joint space yields the cusp point, also the others images of the cusp points will be projections of this fold on the configuration space, where these images correspond to a cusp point formed by the characteristic surfaces. Figure 6 shows that the other images of $CP_{4-6}$ correspond to 3 intersection points between the other images of curves $C_4$ and $C_6$, marked by (*), (**) and (***). As we have 6 cusp points in the section of the joint space represented in figure 5 (the red dots), figure 6 shows 6 tangent points (large red dots) between the DKP singular curves and the curves associated with the characteristic surfaces. Also, the studied cusp $CP_{4-6}$ has 4 distinct DKSs and the other 5 cusps have only 2 distinct DKSs. There are formed by singular curves that separate two regions in the joint space, where one region has 4 DKSs and the other has 2 DKSs. Figure 6 show 3+5=8 cusp points formed by the characteristic surfaces in the workspace (small red dots).

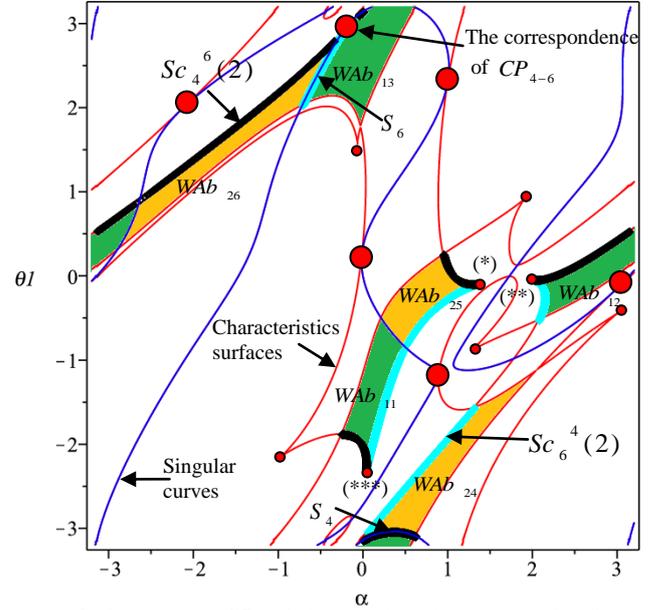

Figure 6. The images of $CP_{4-6}$. Red points show the images of the all cusp points. The largest points are those associated with a tangency point.

### 5.3 Case of a node

In the joint space, a point where two singular curves cross is called a node. Such a point corresponds to the simultaneous coalescence of two couples of DKSs. For the illustrative manipulator, the image in the workspace of a node yields two couples of coincident DKSs lying on the singular curves, plus, possibly, two nonsingular DKSs.

Assume that curve segments $C_h = C_m$ and $C_k = C_n$ intersect in the joint space at node $N_{h-k}$. Since a node comes from the projection of two distinct folds of the configuration space, generically the distribution of DKSs in the four domains when turning around $N_{h-k}$ is $n$, $n-2$, $n-4$ and $n-2$, where $n$ is 4 or 6 for the manipulator under study (Figure 7). Let $WAb_{1h}$, $WAb_{1k}$, $WAb_{2m}$ and $WAb_{2n}$ be the basic regions associated with the four coincident basic components in $QAb_{1h}$, $QAb_{1k}$, $QAb_{2m}$ and $QAb_{2n}$ in the domain with $n \geq 4$ solutions. Note that, $WAb_{1h}$ and $WAb_{1k}$ (resp. $WAb_{2m}$ and $WAb_{2n}$) belong to aspect $WA_1$ (resp. to aspect $WA_2$).

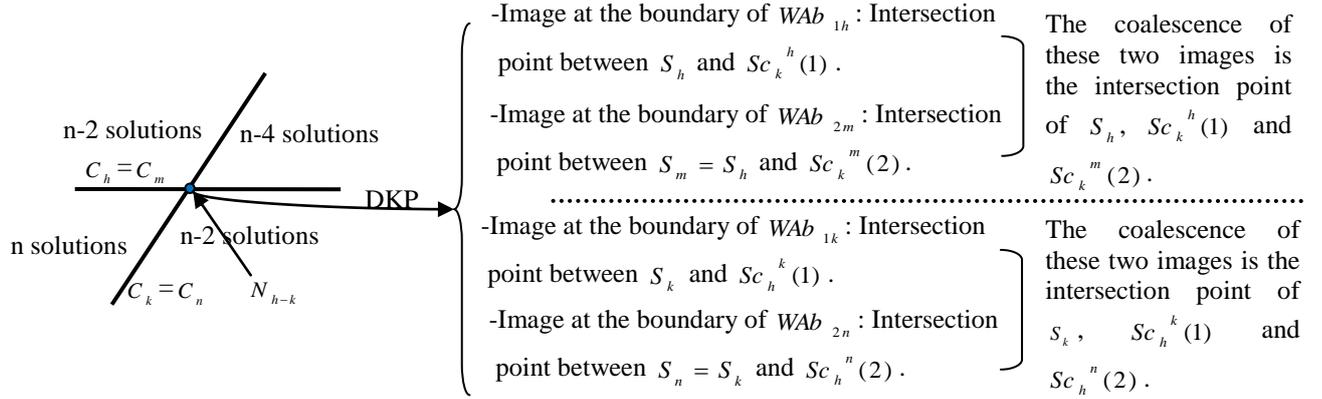

-Image at the boundary of $WAb_{1h}$: Intersection point between $S_h$ and $Sc_k^h(1)$.
-Image at the boundary of $WAb_{2m}$: Intersection point between $S_m = S_h$ and $Sc_k^m(2)$.

The coalescence of these two images is the intersection point of $S_h$, $Sc_k^h(1)$ and $Sc_k^m(2)$.

-Image at the boundary of $WAb_{1k}$: Intersection point between $S_k$ and $Sc_h^k(1)$.
-Image at the boundary of $WAb_{2n}$: Intersection point between $S_n = S_k$ and $Sc_h^n(2)$.

The coalescence of these two images is the intersection point of $S_k$, $Sc_h^k(1)$ and $Sc_h^n(2)$.

Figure 7. The correspondence of the two couples of coincident DKSs of $N_{h-k}$ in the workspace.

Among the images of curve $C_h = C_m$ in the workspace, we have $S_h = S_m$ that defines the singular boundary between basic regions $WAb_{1h}$ and $WAb_{2m}$. Also, $C_h$ yields $Sc_h^k(1)$ and $Sc_h^n(2)$ at the boundaries of $WAb_{1k}$ and $WAb_{2n}$, respectively. Similarly, curve $C_k = C_n$ has one image $S_k = S_n$ at the singular boundary between $WAb_{1k}$ and $WAb_{2n}$, and $Sc_k^h(1)$ and $Sc_k^m(2)$ at the boundaries of $WAb_{1h}$ and $WAb_{2m}$, respectively. Since node $N_{h-k}$ belongs to curves $C_h$ and $C_k$, $N_{h-k}$ has images on all the images of these two curves, especially on $S_h$, $Sc_h^k(1)$, $Sc_h^n(2)$, $S_k$, $Sc_k^h(1)$ and $Sc_k^m(2)$. On the other hand, as $N_{h-k}$ has one image in $WAb_{1h}$ where $S_h$ and $Sc_k^h(1)$ form parts of its boundary, this image is the intersection point between $S_h$ and $Sc_k^h(1)$. By the same way, we can demonstrate that the image of $N_{h-k}$ at the boundary of $WAb_{2m}$ is the intersection point between $S_m$ and $Sc_k^m(2)$. Since these two images of $N_{h-k}$ coalesce at $S_h = S_m$, this coalescence represents the intersection point between $S_h = S_m$, $Sc_k^h(1)$ and $Sc_k^m(2)$ (see figure 7). As $Sc_k^h(1)$ and $Sc_k^m(2)$ lie in two different aspects and represent a part of the curve associated with the characteristic surface, also $S_h$ lies on the singular curves in the workspace, so the first correspondence of a node, representing the coalescence of the first couple of DKSs ($D_h$, $D_m$), is a crossing point between the characteristic surfaces and the singular curves. Similarly, the second image of $N_{h-k}$ representing the coalescence of $D_k$ and $D_n$ at $C_k = C_n$ is the intersection point between $S_k = S_n$, $Sc_h^k(1)$ and $Sc_h^n(2)$ (see figure 7). Therefore, the second correspondence of a node is also a crossing point between the curves associated with the characteristic surfaces and the singular curves.

### 5.4 Example on the node

Node $N_{1-2}$ in figure 5 is formed by curves $C_1$ and $C_2$. Each one of $C_1$ and $C_2$ yields 5 curve images in the workspace, which are colored in green and red, respectively, in figure 8. The curve images colored in blue and light brown are the images of the parts of curves $C_1$ and $C_2$, represented by the same colors in figure 5. $C_1 = C_5$ yields $S_1 = S_5$ at the singular boundary between basic regions $WAb_{11}$ and $WAb_{25}$. Also, $C_1$ yields $Sc_1^2(1)$ and $Sc_1^6(2)$ at the boundaries of $WAb_{12}$ and $WAb_{26}$, respectively. Similarly, curve $C_2 = C_6$ has one image $S_2 = S_6$ at the singular boundary between $WAb_{12}$ and $WAb_{26}$, and $Sc_2^1(1)$ and $Sc_2^5(2)$ at the boundaries of $WAb_{11}$ and $WAb_{25}$, respectively.

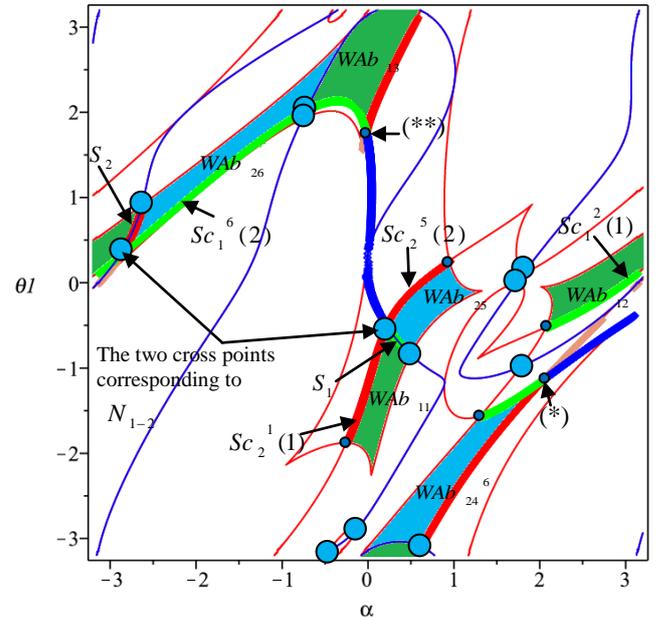

Figure 8. Images of $N_{1-2}$ in the workspace.

Figure 8 shows the two cross points corresponding to $N_{1-2}$; the first one is the intersection point between $S_1$, $Sc_2^1(1)$ and $Sc_2^5(2)$, where the second one is the intersection point between $S_2$, $Sc_1^2(1)$ and $Sc_1^6(2)$.

Note that $N_{1-2}$ has two other image points, which define crossing points between the curves associated with characteristic surfaces. These images are represented by (*) and (**) in figure 8. Due to space limitation, the demonstration is not presented in this paper.

We have 6 double points in the joint space (figure 5), three of them have 4 distinct images and the three remaining ones have 2 distinct images. Figure 8 shows 12 intersection points (the large blue dots) between the singular curves and the curves associated with the characteristic surfaces, and 6 intersection points (the small blue dots) between the curves associated with the characteristic surfaces.

## 6. CONCLUSION

A detailed analysis of singularities in the joint space and in the workspace was presented in this paper. It was shown that the image of a cusp point corresponding to the coalescence of three DKSs, and the image of a double point corresponding to the coalescence of two couples of the DKSs are, respectively, a tangent point and a crossing point between a singular curve and a curve associated with the characteristic surfaces. The demonstration was based on a comprehensive description of the images of the singular curves reflecting the coalescence of the DKSs on it. This preliminary work will help study in detail non-singular assembly-mode changing motions in parallel manipulators.